\pdfoutput=1
\documentclass[letterpaper]{article} 

\usepackage{times,latexsym}
\usepackage{amsfonts}
\usepackage{comment}
\usepackage{amsmath}
\usepackage{subcaption} 
\usepackage[T1]{fontenc}
\usepackage{enumitem}
\usepackage{booktabs}
\usepackage{makecell}

\usepackage{aaai21}  
\usepackage{times}  
\usepackage{helvet} 
\usepackage{courier}  
\usepackage[switch]{lineno} 
\usepackage[hyphens]{url}  
\usepackage{graphicx} 
\usepackage{natbib}  
\usepackage{caption} 
\title{InSRL: A Multi-view Learning Framework Fusing Multiple Information Sources for Distantly-supervised Relation Extraction}

\author {
        Zhendong Chu\textsuperscript{\rm 1}\footnote{equal contribution}, 
        Haiyun Jiang\textsuperscript{\rm 2}\footnotemark[1], 
        Yanghua Xiao\textsuperscript{\rm 3},
        Wei Wang\textsuperscript{\rm 3}\\
}
\affiliations {
    \textsuperscript{\rm 1}University of Virginia, \textsuperscript{\rm 2}Tencent AI Lab, \textsuperscript{\rm 3}Fudan University \\
    zc9uy@virginia.edu, haiyunjiang@tencent.com,  \{shawyh, weiwang1\}@fudan.edu.cn
}

\begin{document}
\maketitle

\begin{abstract}
Distant supervision makes it possible to automatically label bags of sentences for relation extraction by leveraging knowledge bases, but suffers from the \emph{sparse} and \emph{noisy bag} issues. Additional information sources are urgently needed to supplement the training data and overcome these issues. In this paper, we introduce two widely-existing sources in knowledge bases, namely entity descriptions, and multi-grained entity types to enrich the distantly supervised data. We see information sources as multiple views and fusing them to construct an intact space with sufficient information. An end-to-end multi-view learning framework is proposed for relation extraction via Intact Space Representation Learning (InSRL), and the representations of single views are jointly learned simultaneously. Moreover, inner-view and cross-view attention mechanisms are used to highlight important information on different levels on an entity-pair basis. The experimental results on a popular benchmark dataset demonstrate the necessity of additional information sources and the effectiveness of our framework. We will release the implementation of our model and dataset with multiple information sources after the anonymized review phase.
\end{abstract}

\section{Introduction}
\pdfoutput=1
Relation extraction (RE) plays a vital role in natural language understanding, knowledge graph construction and other important natural language processing tasks \cite{etzioni2004web, lin2016neural}. It aims to identify the semantic relations between an entity pair in the given sentences. The bottleneck of the learning-based RE solutions is the lack of large-scale manually labeled data. Based on a simple assumption that, for a triplet  <$e_1, r, e_2$>  existing in Knowledge Base (KB), any sentence mentions the two entities might express the relation $r$, Distant Supervision (DS) \cite{mintz2009distant} is proposed to automatically construct training data.

However, data generated by distant supervision contains too much noise to train an effective RE model. Many existing solutions \cite{ji2017distant, zeng2014relation, yuan2019cross} construct sentence bags and employ Multi-instance Learning (MIL) for bag-level RE, which is based on the ``at-least-one'' assumption, i.e., at least one sentence in a bag clearly expresses the target relation \cite{hoffmann2011knowledge}. Nonetheless, this assumption is too strong to be valid because of two facts in real datasets. The first is that most entity pairs have only \emph{one} labeled sentence (81.9\% in the training set of NYT dataset), which leads to the \emph{sparse bag} problem and makes the setting of MIL meaningless.
The second is, for some entity pairs, all the sentences in the bag do not express the relations, i.e., the \emph{noisy bag} problem. The problem is very common in the datasets generated by distant supervision. For example, up to 53\% of 100 randomly sampled bags are noisy in NYT dataset \cite{feng2018reinforcement}. Thus, it is hard to get sufficient information for the RE task only from the DS-labeled sentences.

\begin{figure*}[htbp]
	\centering
	\setlength{\belowcaptionskip}{-12pt}
	\includegraphics[scale=0.57]{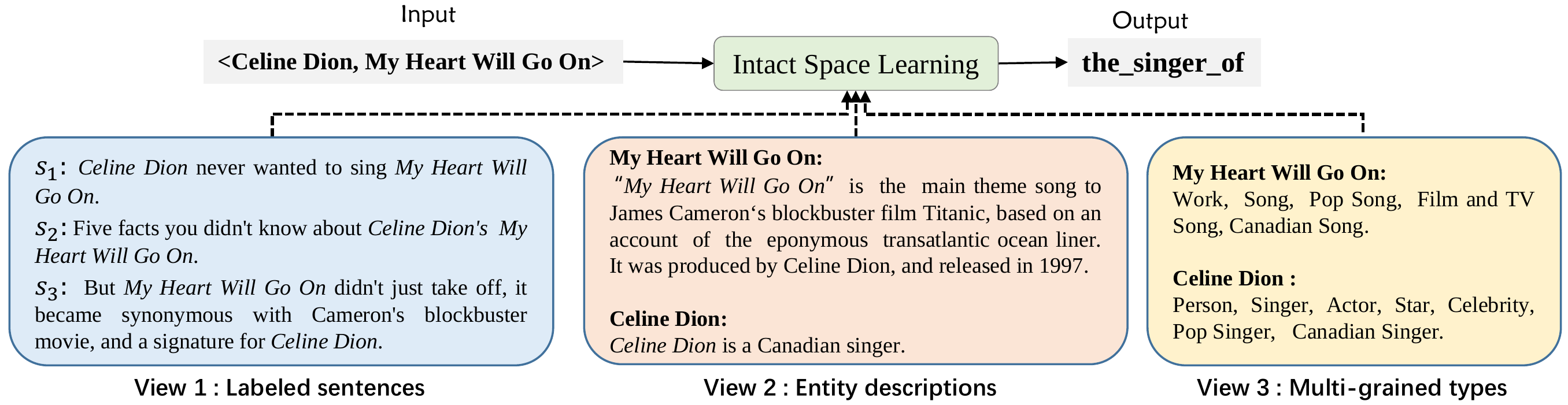}
	\caption{An example of inferring the target relation by utilizing multiple information sources.}
	\label{fig:multiview_example}
\end{figure*}

One way to alleviate the aforementioned two issues of DS is fusing additional sources of information with labeled sentences. In this paper, we introduce two widely-existing sources in KB, namely \emph{entity descriptions} and \emph{multi-grained entity types}, as complements to labeled sentences. Entity descriptions contain the summary information of entities, thus providing rich background knowledge for relation identification \cite{xie2016representationDescriptions,ji2017distant, zhong2015aligning}. Besides, multi-grained types usually provide rich constraint information for RE \cite{ling2012fine,xie2016representation123, han2018hierarchical}.
For example in Figure \ref{fig:multiview_example}, each labeled sentence fails to directly express the target relation. Thus, it is difficult for machines to accurately determine the target relation under the setting of MIL. However, the description of $[\emph{Celine Dion}]_{e_1}$ indicates she is a singer and $[\emph{\text{My Heart Will Go On}}]_{e_2}$ is a song. The entity types provide further verification. Hence, the target relation \textbf{\emph{the\_singer\_of}} can be easily inferred. Therefore, fusing multiple information sources make it possible to learn more informative representations for the RE task. 

However, it is not trivial to effectively fuse information from multiple sources. Multi-view learning (MVL) has been proved to be a powerful way to learn from multiple sources \cite{blum1998combining,sindhwani2005co,xu2013survey,xu2015multi}. In the context of RE, every information source of an entity pair is regarded as a view, all of them could express the target relation. By fusing different individual views, we eventually learn a more informative representation for RE. However, most of the MVL algorithms, e.g., co-training \cite{blum1998combining} or co-regularization \cite{sindhwani2005co}, require the prerequisite of \emph{view sufficiency}.
That is, the information of each view (source) needs to be sufficient enough to learn a high-quality representation, which does not hold in our setting since each view only contains \emph{partial} information due to the sparse and noisy issues, in other words, view sufficiency is not satisfied in DS-based RE. This motivates us to use the idea of \emph{intact space learning} (ISL) \cite{xu2015classifying, lin2017multi, huang2019multi}, where view sufficiency assumption is not needed. The basic assumption is there exists an intact space and each view is an insufficient projection of it. Even individual view only captures partial information of an entity pair, we can obtain redundant information from its intact space representation by integrating them. To the best of our knowledge, this is the first attempt applying MVL to RE and realize it with more practical assumptions.

However, the standard ISL framework is built upon the two-stage strategy where single view and intact space representations are learned separately, which is with poor efficiency and hard to tune. In this paper, we propose an end-to-end framework to jointly learn single view and intact space representations for RE. Additionally, inner-view and cross-view attention mechanisms are well-designed for our ISL-based RE solution. Inner-view attention ensures that relation-aware features can be captured in each view, while cross-view attention is proposed to highlight important views when constructing the intact space on an entity-pair basis (e.g., entity description contains more important information for \textbf{\emph{the\_singer\_of}}  in Figure \ref{fig:multiview_example}). At last, a relation classifier is trained upon the intact space representations. We name this framework as \textbf{\underline{In}}tact \textbf{\underline{S}}pace \textbf{\underline{R}}epresentation \textbf{\underline{L}}earning (InSRL). Extensive experiments demonstrate the necessity of additional information sources and the effectiveness of our framework, showing significant improvement comparing to state-of-the-art approaches.

\section{Method Overview}

\subsubsection{Notations and Problem Definition} We denote a training sample\footnote{Note that some entity pairs may have multiple relations in the dataset. For these entity pairs, we generate multiple samples and each sample only involves one different relation.} as $(v_1, v_2, v_3; r)$, where $v_1,v_2,v_3$ are the input data of the entity pair $t = \{e_1, e_2\}$, $e_1$ is the head entity, $e_2$ is the tail entity, and $r \in \mathcal{R}$ is the target relation.
In particular, $v_1=\{s_1,...,s_n\}$ is the bag of labeled sentences that contains the entity pair.
$v_2=\{d_1,d_2\}$ denotes the entity descriptions of corresponding entities.
$v_3 = \{\mathcal{C}_1, \mathcal{C}_2\}$, where $\mathcal{C}_1$ and $\mathcal{C}_2$ are the multi-grained type sets of $e_1$ and $e_2$, respectively.
Our goal is to learn a mapping $r=f(v_1,v_2,v_3)$ to predict the target relations for unseen entity pairs.

\begin{figure*}[htbp]
\begin{subfigure}{.6\textwidth}
  \centering
  \includegraphics[height=8.5cm]{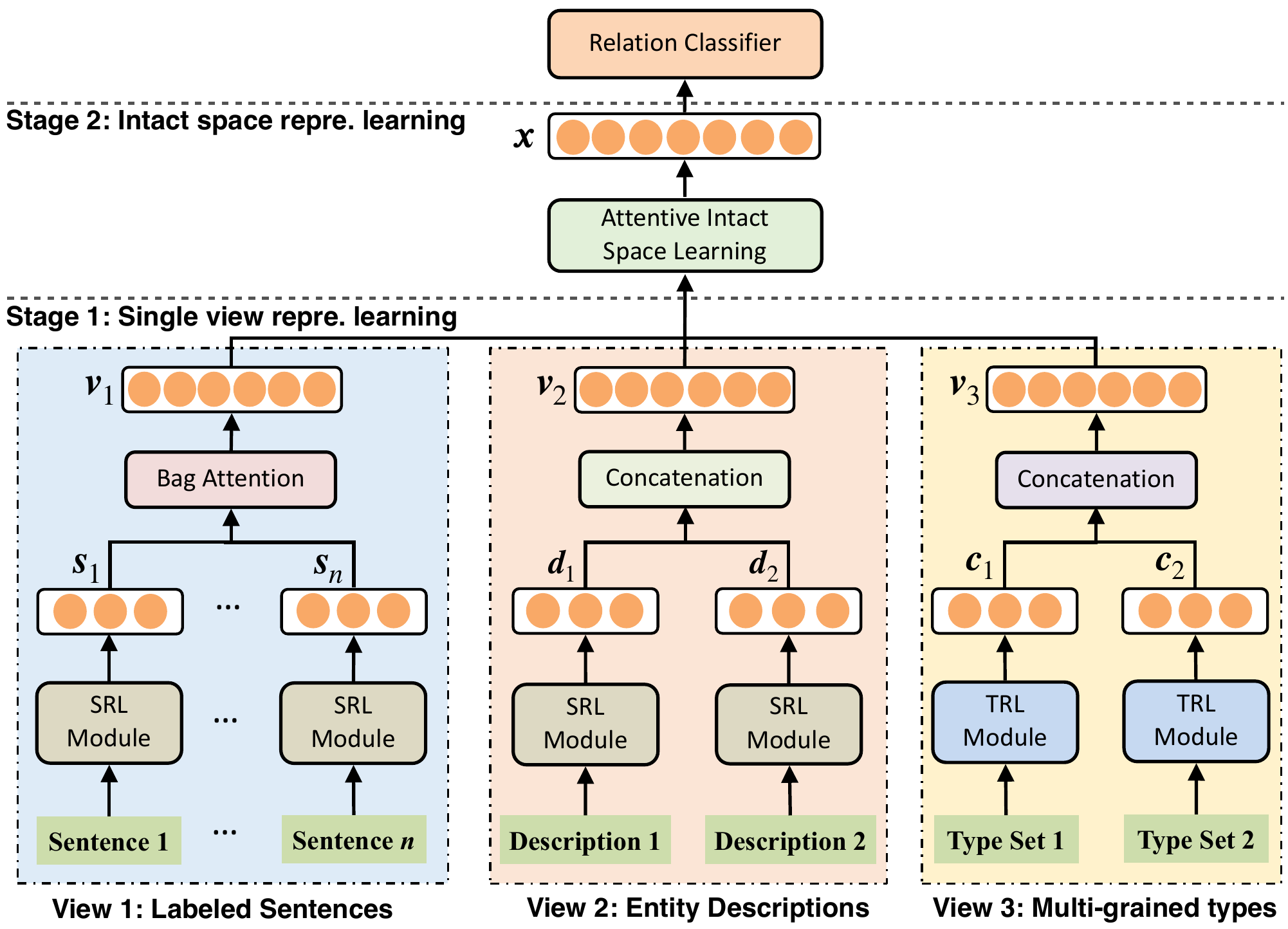}
  \caption{}
  \label{fig:framework}
\end{subfigure}
\begin{subfigure}{.53\textwidth}
  \centering
  \includegraphics[height=8.5cm]{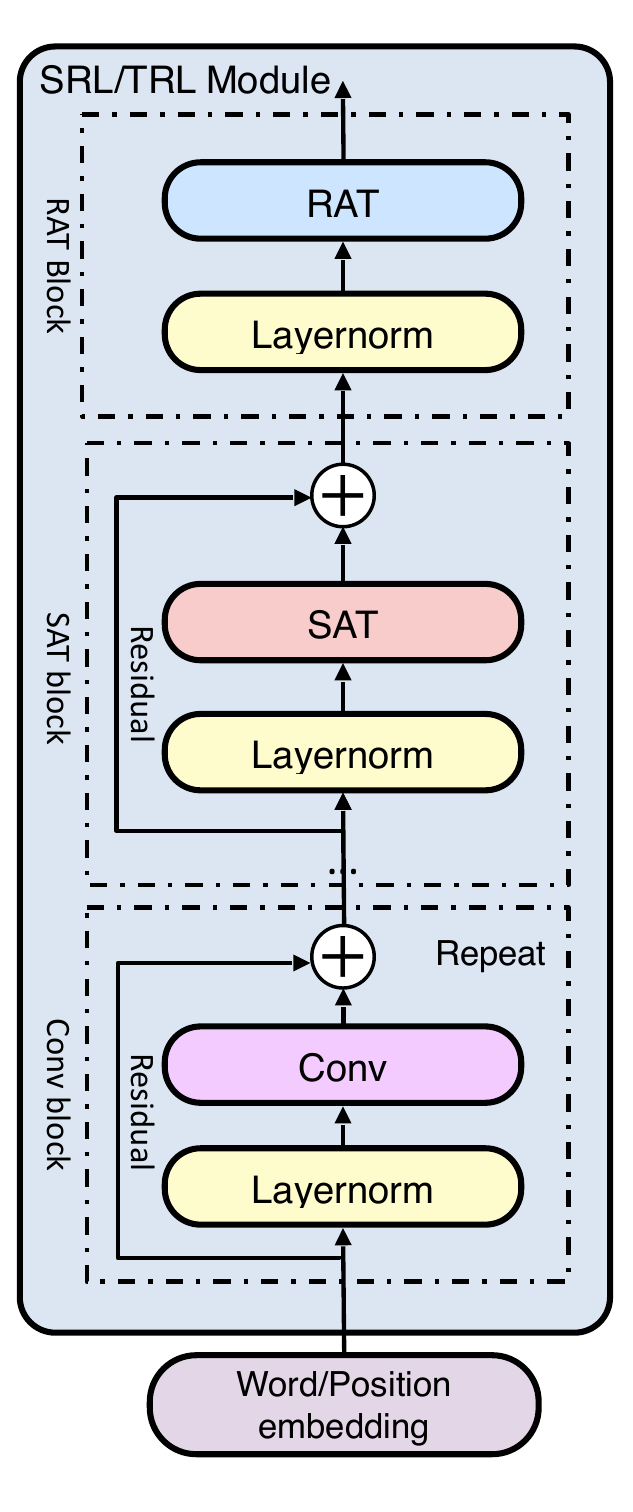}
  \caption{}
  \label{fig:srl}
\end{subfigure}
\caption{(a) The architecture of InSRL framework. (b) The structure of SRL/TRL module, convolutional blocks are removed when learning type representations.}
\end{figure*}

\subsubsection{Model Components}As shown in Figure \ref{fig:framework}, InSRL mainly contains three components: (1) single view  representation learning, (2) attentive intact space representation learning, (3) relation classifier training. In the first stage, we aim to obtain relation-sensitive single view representations $\{\boldsymbol{v}_1, \boldsymbol{v}_2, \boldsymbol{v}_3\}$. To this end, we design novel transformer-like modules named SRL/TRL (SRL for sentence representation learning, TRL for type representation learning) with the structure in Figure \ref{fig:srl}. The difference between them is convolutional blocks are set to capture local information of sentences in SRL. Besides, we propose a novel inner-view relation-aware attention block (i.e., RAT) to highlight task-specific features for the RE task. The learned single view representations are used to construct the intact space representations in the second stage. In particular, we design a cross-view attention mechanism to place emphasis on more important views on an entity-pair basis. Finally, the relation classifier $r=h(\boldsymbol x)$ can be easily built by taking the intact space representations as inputs. We will first deliver the details of the single view representation learning part in the next section, and then describe how to learn intact space representations and train the relation classifier.

\section{Single View Representation Learning} \label{sec:single_view}






The structure of SRL/TRL is shown in Figure \ref{fig:srl}, which contains three blocks: convolutional (Conv), self-attention (SAT), and relation-aware attention (RAT). The Conv block extracts the local features of a sequence while the SAT block captures the long-term dependencies in the sequence.
Both two blocks contain the residual connection and layer normalization \cite{ba2016layer}. In general, the two tricks make the proposed model more easy and stable in optimization \cite{he2016deep,vaswani2017attention}. The Conv block is removed when learning type set embeddings since the order in the type set is meaningless. Furthermore, the RAT block is proposed to extract relation-sensitive features from the outputs of the former two blocks. This block also contains the layer normalization.
Next, we first describe the inputs of SRL/TRL modules and then detail the blocks.

\subsubsection{Input embeddings} The inputs of the SRL module contains word and position embeddings. Given a sentence $s$ containing $l$ words\footnote{The length of $s$ is either truncated or padded to $l=120$ in this paper.}, the embedding of each word $w_i$ is denoted as $\boldsymbol{e} = [\boldsymbol{e}_w;\boldsymbol{e}_p] \in \mathbb{R}^{d_a+d_b}$, where $\boldsymbol{e}_w \in \mathbb{R}^{d_a}$ and $\boldsymbol{e}_p \in \mathbb{R}^{d_b}$ are the word and position embeddings, respectively.
The word embeddings are initialized by word2vec \cite{mikolov2013distributed}, which will be described in the experimental section.

The position embeddings help the model to keep track of how close each word is to the head or tail entity. Besides, they can also capture the order information in a sentence.
Given a labeled sentence, the position embedding of each word is derived following \cite{zeng2015distant}.
Specifically, we allocate two different dense vectors with dimension $d_b/2$ to encode the relative distances from the current word $w_i$ to the head or tail entities.
Then we concatenate them into one vector with dimension $d_b$, i.e., $\boldsymbol{e}_p$.
Note that we keep another set of position embeddings for entity descriptions, because we only need to encode the relative distance from the current word $w_i$ to the corresponding entity into one vector, so the dimension is $d_b$.
All the position embeddings are randomly initialized.
Both the position and word embeddings will be fine-tuned during the training process. Moreover, the dimension of $\boldsymbol{e}$ is first transformed to $d_{m}$ for the later relation-aware attention calculation, with the help of a linear matrix $\boldsymbol{M} \in \mathbb{R}^{d_{m} \times (d_a+d_b)}$, i.e., $\boldsymbol{e}' = \boldsymbol{M}\boldsymbol{e}\in\mathbb{R}^{d_{m}}$.
$d_{m}=128$ is used throughout the model.
Since each $s$ contains $l$ words, we denote the embedding matrix of $s$ as $\boldsymbol{E}' \in  \mathbb{R}^{{d_{m}} \times l}$.

The inputs of the TRL module are embeddings of types in a type set $\mathcal{C}$. For the convenience of implementation, we fix the size of $\mathcal C$ to $l_0=15$ by randomly discarding some types or padding with ``null".
For each type $c \in \mathcal C$ (including ``null''), the embedding is randomly initialized as $\boldsymbol c \in \mathbb{R}^ {d_c}$, where $d_c$ is set to 16.
Note that the type set is unordered, thus position embeddings are not needed. Similarly, we transform $\boldsymbol{c}$ into a vector $\boldsymbol{c}'$ dimension of $\boldsymbol c$ to $d_{m}$ with a parameter matrix $\boldsymbol{M}_1 \in\mathbb{R}^{d_{m}\times d_c}$, i.e., $\boldsymbol { c}'= \boldsymbol M_1\boldsymbol c$.
Since the type set $\mathcal C$ contains $l_0$ types, we denote the input embedding matrix of $\mathcal C$ as $\boldsymbol C' \in \mathbb{R}^{d_{m} \times l_0}$. 

The obtained embeddings are inputed into the following blocks to extract local features and dependenies among words/types.

\subsubsection{Convolutional (Conv) and Self-attention (SAT) Blocks}We follow the design of the convolutional and self-attention blocks in \cite{yu2018qanet}.
In the Conv block, the size of convolutional kernels is 7 and the filter number is 128.
The Conv block contains 4 repeating convolutional layers. The input of this block is the embedding matrix  of $s$, i.e., $\boldsymbol{E}'$ and the output is also a matrix with dimensions ${d_{m} \times l}$. In the self-attention block, the multi-head attention \cite{vaswani2017attention} is adopted with 8 heads.
The input and output of this block are also matrices with dimensions ${d_{m} \times l}$.
More details can be found in \cite{yu2018qanet}. When learning the representations of type sets, we remove the Conv block, and directly input the type embeddings into the SAT block.

\subsubsection{Relation-aware Attention (RAT) Block}This block highlights the relation-sensitive features in the outputs of former blocks. 
We denote the outputs of Layernorm($\cdot$) in this block as $\boldsymbol{H} \in \mathbb{R}^{d_{m}\times l}$.
Then the output of this block $\boldsymbol{o} \in \mathbb{R}^{d_{m}}$ is computed as:
\begin{equation}
\begin{gathered}
    \boldsymbol{o} = \boldsymbol{H}\boldsymbol{\alpha}   \\
\boldsymbol{\alpha} = \text{softmax}([\alpha_{1},\alpha_{2},...,\alpha_{l}])^T \\
\alpha_{i} =   \boldsymbol{w}^T \tanh(\boldsymbol{W} \boldsymbol{h}_i + \boldsymbol{\hat r})+b
\end{gathered}
\end{equation}
where $\alpha_{i}$ is the attention weight of the RE task to the $i$-th column in $\boldsymbol{H}$ (i.e., $\boldsymbol{h}_i$).
$\boldsymbol{w}\in\mathbb{R}^{d_{m}}, \boldsymbol{W} \in \mathbb{R}^{d_{m} \times d_{m}}$ and $b \in \mathbb{R}$ are parameters.
$\boldsymbol{\hat r}$ is a task-related \emph{query} vector that is defined as:
\begin{equation}
\boldsymbol{\hat r} = \frac{1}{n} \sum_{r_k \in \mathcal R}  \boldsymbol r_k
\label{global relation embedding}
\end{equation}
where $\boldsymbol r_k \in \mathbb{R}^{d_{m}}$ is the embedding of the  relation $r_k$ and $n = |\mathcal R|$ is the relation size.
That is, $\boldsymbol{\hat r}$ captures the global relation information of the RE task. The relation embeddings will also be used to calculate the importance of different views later. In our model, all the relation embeddings will be learned with other parameters and be shared across all the modules during the training process. The shared relation embeddings help us learn more task-specific representations.

\vspace{0.3cm}

\subsection{View 1 : Labeled Sentences}
We model the learning of a labeled sentence bag as a multi-instance learning problem \cite{lin2016neural,ji2017distant,jiang2016relation}. The importance of sentences are various due to its similarity with a target relation, and highlighting the important sentence will relieve the noisy bag problem. To suppress noise, we follow the selective attention in \cite{lin2016neural,ji2017distant} takes the sentence-level attention, where an importance weight is learned for each sentence.  Specifically, for each sentence $s_i \in v_1$, we take the SRL model to extract relation-sensitive features, which is denoted as $\boldsymbol{s}_i \in \mathbb{R}^{d_{m}}$.
Based on the selective attention, the representation of the sentence bag (denoted as $\boldsymbol{v}_1 \in \mathbb{R}^{d_{m}}$) is computed as
\begin{equation}
\boldsymbol{v}_{1} = \sum\nolimits_{i=1}^{m}{\beta_{i} \boldsymbol{s}_{i}},\\
\end{equation}
where $\beta_i$ is the attention weight, which is computed by,
\begin{equation}\label{eq1}
\beta_{i} = \frac{\text{exp}(\omega_i)}{\sum_{j=1}^m\text{exp}(\omega_j)},
\end{equation}
where $\omega_i$ refers to the scores indicating how well the input sentence $s_i$ and the predict relation $r$ matches, which is computed by,
\begin{equation}
    \omega_i = \boldsymbol{s}_i \boldsymbol{W}_1 \boldsymbol{r}.
\end{equation}

where $\boldsymbol{W}_1$ is a weighted diagonal matrix, and $\boldsymbol{r}$ is the query vector associated with relation $r$ which indicates the relation representation shared across the model.

\subsection{View 2 : Entity Descriptions}
Entity descriptions contain abundant background knowledge for RE task.
For example, the descriptions of  \texttt{Celine Dion} and \texttt{My Heart Will Go On} are very helpful to identify the target relation \emph{the\_singer\_of} (shown in Figure \ref{fig:multiview_example}), because there are many key signals, e.g.,  ``theme song'', ``recorded by Celine Dion'', ``Canadian singer''.

We describe how to extract informative features from the source of entity descriptions.
For each entity description $d_1$ or $d_2$, we take the SRL module to derive the relation-sensitive representation and denote the outputs as $\boldsymbol{d}_1$ and $\boldsymbol{d}_2$, respectively.
Then we concatenate them into $[\boldsymbol{d}_1;\boldsymbol{d}_2] \in \mathbb{R}^{2d_{m}}$ and take a nonlinear function to obtain the final view representation $\boldsymbol v_2 \in \mathbb{R}^{d_{m} }$.
That is, 
\begin{equation}
\boldsymbol v_2 =  \tanh (\boldsymbol W_2 [\boldsymbol{d}_1;\boldsymbol{d}_2]+\boldsymbol b_2)
\label{obtain the final view representation v2}
\end{equation}
where $\boldsymbol W_2 \in \mathbb{R}^{d_{m} \times 2d_{m}}$ reduces the dimension to $d_{m}$ and $\boldsymbol b_2 \in \mathbb{R}^{d_{m}}$ is a bias. The nonlinear function also ensures the consistency of the dimensions of different single view embeddings.

\subsection{View 3 : Multi-grained Types}
Multi-grained types contain rich indicative information for relation identification.
For example, the types ``Singer'' and ``Song'' for the entities \texttt{Celine Dion} and \texttt{My Heart Will Go On} indicates the relation \emph{the\_singer\_of} may hold between this entity pair.
Furthermore,  different types will have a distinguishable contribution to the relation inference.
For example, the type ``Pop Song'' of \texttt{Celine Dion}  is more important than ``Person" for the inference of \emph{the\_singer\_of}. 
The unbalanced contributions of types can be well captured by the self-attention mechanism.

We describe how to obtain the view embedding $\boldsymbol v_3 $ given the multi-grained types $(\mathcal C_1, \mathcal C_2)$.
For each type set $\mathcal C_1$ or $\mathcal C_2$, we take the TRL module to extract the relation-sensitive features and denote them as $\boldsymbol c_1, \boldsymbol c_2 \in \mathbb{R}^{d_{m}}$, respectively.
Similar to the learning of entity description pairs, we also concatenate $\boldsymbol c_1, \boldsymbol c_2$ following a nonlinear transformation, i.e., 
\begin{equation}
\boldsymbol v_3 =  \tanh (\boldsymbol W_3 [\boldsymbol{c}_1;\boldsymbol{c}_2]+\boldsymbol b_3)
\label{obtain the final view representation v3}
\end{equation}
where $\boldsymbol W_3 \in \mathbb{R}^{d_{m} \times 2d_{m}}$ and $\boldsymbol b_3 \in \mathbb{R}^{d_{m}}$ are parameters.




\section{Intact Space Representation Learning \\ \& End-to-end Training Strategy}\label{sec:intact_space}
\subsection{Intact Space Learning}
We have obtained the single view feature vectors $\boldsymbol{v}_1$,  $\boldsymbol{v}_2$ and $\boldsymbol{v}_3$ above.
Since every single view is insufficient to train a high-performance and robust relation classifier, we hope to obtain a sufficient embedding for the entity pair by integrating the three views.
In this paper, the idea of intact space learning \cite{xu2015multi} is exploited.
The basic assumption is that, given an entity pair, the feature vector of each view is generated by a latent representation in an intact space and this representation is sufficient for RE task.
Thus, our goal is to obtain the latent representation $\boldsymbol x$ for an entity pair $t$ based on $\boldsymbol{v}_1$,  $\boldsymbol{v}_2$ and  $\boldsymbol{v}_3$.
The theoretical analysis of intact space learning within this assumption has been presented in \cite{xu2015multi}, which proves that the complementarity between multiple views is beneficial for the stability and generalization.


Specifically, we assume that $\boldsymbol{v}_j$ ($j=1,2,3$) is generated by $\boldsymbol{x} \in \mathbb{R}^{d_{x}}$ with the help of a view generation function $f_j$, that is,  
\begin{equation}
\boldsymbol{v}_j = f_j(\boldsymbol{x}) + \boldsymbol{\epsilon}_j,  \qquad j=1,2,3
\end{equation}
$\boldsymbol{\epsilon}_j$ is the view-dependent noise introduced by the single view learning module.
For computational effectiveness, we approximate $f_j$ with  a linear function, i.e., $f_j(\boldsymbol{x}) = \boldsymbol{\hat{W}}_j \boldsymbol{x}$, where  $\boldsymbol{\hat{W}}_j \in \mathbb{R}^{d_{m} \times d_x}$ and  $d_{x}>d_{m}$.
To obtain $\boldsymbol x$, a straightforward approach is to minimize the empirical risk over $\{\boldsymbol v_j-f_j(\boldsymbol{x})\}^3_{j=1}$.
Furthermore, each view provides an \emph{unbalanced  contribution} to the relation identification.
Thus, the learning of $\boldsymbol{x}$ should pay more attention to the important views.
However, this principle is ignored in the early intact space learning. 
In this paper, we conduct the \emph{view-level attention} to learn the importance weight of each view.
The loss function to learn $\boldsymbol x$ is defined as 
\begin{equation}
\text{loss}(t)= 
\sum_{j=1}^{3} \gamma_{j} ||\boldsymbol{v}_j-\boldsymbol{\hat{W}}_j \boldsymbol{x}||^2 
\label{loss function single sample}
\end{equation}
where $\boldsymbol{\epsilon}_j = \boldsymbol{v}_j-  \boldsymbol{\hat{W}}_j \boldsymbol{x}$ and $||\cdot||$ is $L_2$ norm.
The view-level attention is considered.
$\gamma_j$ ($j=1,2,3$) is the attention weight of the RE task to view $\boldsymbol v_j$, which is computed as
\begin{equation}
\begin{gathered}
\gamma'_j = \boldsymbol{w}_4^T \tanh(\boldsymbol{W}_4 \boldsymbol{v}_j + \boldsymbol{\hat r})+b_4 \\
\gamma_j = \frac{\exp(\gamma'_j)}{\sum_{k=1}^{3}  \exp(\gamma'_k)}
\end{gathered}
\label{attention weight of the RC task to view}
\end{equation}
where $\boldsymbol{w}_5 \in \mathbb{R}^{d_m}$, $\boldsymbol{W}_4\in  \mathbb{R}^{d_m \times d_m}$ and $b_4$ are parameters.
$\boldsymbol{\hat r}$ is the query vector.

When considering all the samples in the training dataset, the loss function for obtaining latent embeddings is defined as   
\begin{equation}
\mathcal L_{I} ({\varTheta})= 
\mathop   \frac{1}{D} \sum_{i=1}^D  \text{loss}(t_i)
\label{is extened as }
\end{equation}
where $t_i$ is the entity pair in the $i$-th sample and $D$ is the sample size.
Note that the features $\boldsymbol v_1, \boldsymbol v_2$ and $\boldsymbol v_3$ are derived by deep modules that also need to be learned.
In this paper, we \emph{jointly learn the single view and intact space representations}.
Thus, the parameters $ \varTheta$ come from two aspects: single view learning and intact space learning.
The parameters in intact space learning consist of the generation matrices $\{\boldsymbol{\hat{W}}_j\}$ ($j=1,2,3$), the latent embeddings $\{\boldsymbol x_i\}$ ($i=1,...,D$) and the parameters in Eq \eqref{attention weight of the RC task to view}.
In other words, the optimization of $\mathcal L_{I} ({\varTheta})$ will generate the latent embeddings for entity pairs in the training dataset.

\subsection{Relation Classifier}
We construct the relation classifier with the latent embeddings as inputs.
Given the latent embedding $\boldsymbol{x}_i$ of an entity pair $t_i$, we define the conditional probability $p(r_i|\boldsymbol{x}_i)$  with a  Softmax layer, that is,
\begin{equation}
p(r_i|\boldsymbol{x}_i) =   \frac{\exp(\boldsymbol{r_i}^T\boldsymbol{M}_2\boldsymbol{x}_i)}{\sum_{k=1}^{n} \exp(\boldsymbol{r}_k^T\boldsymbol{M}_2\boldsymbol{x}_i)}
\label{Softmax layer as follows}
\end{equation}
$\boldsymbol R=[\boldsymbol r_1;\boldsymbol r_2;...;\boldsymbol r_n] \in \mathbb{R}^{d_{m}\times n}$ is the embedding matrix of the relations to be learned.
Note that $\boldsymbol R$ are also the parameters in attention operations, which aims to derive the query embedding $\boldsymbol {\hat r}$ (first defined in Eq  \eqref{global relation embedding}).
$\boldsymbol{M}_2\in\mathbb{R}^{d_{m} \times d_x}$ is a weighted matrix for the similarity computation between $\boldsymbol{r}_k$ and $\boldsymbol x$.
We denote ${\Phi} = \{\boldsymbol R, \boldsymbol M_2\}$.
Then the loss function for learning relation classifier is defined using cross-entropy, i.e., 
\begin{equation}
\mathcal L_{RC}({\Phi}) = 
-\frac{1}{D} \sum_{i=1}^D \log p(r_{i}|\boldsymbol{x}_i)
\label{function for RC task is}
\end{equation}
where $\boldsymbol{x}_i$ and $r_{i}$ are the latent embedding and target relation of $t_i$, respectively.

\subsection{End-to-end training strategy}
In traditional ISL framework, only the latent space representations are needed to learn. However, in our setting, we also need to learn all the view representations as well as the latent feature vector for each entity pair, which makes our model learning challenging.
Specifically, our framework contains two loss functions: $\mathcal L_I ({\varTheta})$ and $\mathcal L_{RC}({\Phi})$.
The former aims to learn the features of each view as well as the intact space embeddings and the latter aims to train the relation classifier. 
To learn all the parameters, the direct solution is the \emph{sequential} training, which is adopted in \cite{xu2015multi}.
That is, we first minimize $\mathcal L_I ({\varTheta})$ and obtain the latent embeddings $\{\boldsymbol{x_i}\}$.
Then we train the relation classifier using $\{\boldsymbol{x}_i\}$ as inputs, i.e., minimizing $\mathcal L_{RC}({\Phi})$, where the relation embeddings $\boldsymbol R$ are initialized by the results in the first step.
However, this training strategy is not effective and sufficient because the learned intact space embeddings are \emph{not sensitive} to the RE task itself. In other words, intact space representation learning is not guided by the optimization of RE performance.

Hence, we propose to jointly optimize $\mathcal L_I ( {\varTheta})$ and $\mathcal L_{RC}( {\Phi})$ in an end-to-end manner.
In $\mathcal L_I ( {\varTheta})$, the latent representations $\{\boldsymbol x_i\}$ ($i=1,...,D$) are parts of the parameters to be learned.
We minimize $\mathcal L_I ( {\varTheta})$ by setting the gradient of the function with respect to $\boldsymbol x_i$ to 0, 
\begin{equation}
\sum_{j=1}^{3} \gamma_{j} \boldsymbol W^T_j(\boldsymbol v_j - \boldsymbol W_j \boldsymbol x_i) = 0
\label{}
\end{equation} 
This will produce the closed-form solution for $\boldsymbol x_i$.
That is, 
\begin{equation}
\boldsymbol x_i = \left(\sum_{j=1}^{3} \gamma_{j} \boldsymbol W^T_j \boldsymbol W_j \right)^{-1} \sum_{j=1}^{3} \gamma_{j} \boldsymbol W^T_j \boldsymbol v_j
\label{eq:closed-form}
\end{equation} 
We conclude that $\boldsymbol x_i$ is the function of all the parameters in $\mathcal L_I ( {\varTheta})$\footnote{The calculation of the inverse matrix is inefficient in practice, we replace it with a fully-connected layer in our implementation. The formulation becomes $\boldsymbol x_i = \boldsymbol W \sum_{j=1}^{3} \gamma_{j} \boldsymbol W^T_j \boldsymbol v_j$, where $\boldsymbol W$ is a learnable weight matrix.}.
As a result, we take the expression of $\boldsymbol x_i$ into Eq  \eqref{Softmax layer as follows}, and build the relationship between $\mathcal L_I ( {\varTheta})$ and $\mathcal L_{RC}( {\Phi})$.
In this way, we only need to optimize $\mathcal L_{RC}( {\Phi})$ and it will optimize the parameters in $\mathcal L_{RC}( {\Phi})$ and $\mathcal L_I ( {\varTheta})$ simultaneously (including $\{\boldsymbol x_i\}$ ($i=1,...,D$)).

\section{Experiments}
In this section, we first introduce the dataset and evaluation metrics. We also present the detailed experimental settings. Then, we compare our model with several state-of-the-art methods. At last, we compare our attentive intact space learning methods with other information fusion methods to show its effectiveness.


\subsubsection{Dataset and Evaluation Metrics}The widely used NYT \cite{Riedel2010Modeling} dataset is considered.
There are 522,611 labeled sentences, 281,270 entity pairs, and 18,252 relational facts in the training set; and 172,448 sentences, 96,678 entity
pairs and 1,950 relational facts in the test set. 
The dataset contains 53 unique relations including a special relation ``NA'' that denotes no relation between two entities. There are a total of  39,529 entities in  NYT, and 25,271 of them  have unique descriptions in Freebase\footnote{https://developers.google.com/freebase}. The multi-grained types for an entity are also obtained from Freebase. Every entity has 7.83 entity types on average.

We follow the evaluation metrics in previous works \cite{ye2019distant, vashishth-etal-2018-reside}, where precision-recall curves, AUC and max F1 scores on the held-out set are presented. The relations extracted from the held-out set are compared with those in Freebase, which save the costly human evaluation.

\subsubsection{Experimental Settings}We use word2vec\footnote{https://code.google.com/p/word2vec/} to initialize the word embeddings on the combination of the NYT corpus and the entity description set with dimension $d_a = 50$.
We set the dimensions of position embedding $d_b = 14$ and type embedding $d_c=16$. The dimensions of intact space embeddings $d_{x}=\{300,350,400, 450, 500\}$ are considered, respectively. We set $d_{x}=400$ by default, which achieves best performance in practice.

We train our model on a GTX 1080Ti GPU with 10GB graphic memory.
We implement our framework using Pytorch. We choose SGD as our optimizer. 
The batch size is 200 and the learning rate is 0.01.
Besides, we run our model for 5 runs with different random seeds,  each with 80 iterations, and then we report the average performance with the standard error. 

\begin{figure*}
\centering
\begin{subfigure}{0.45\textwidth}
\centering
    \includegraphics[width=7cm]{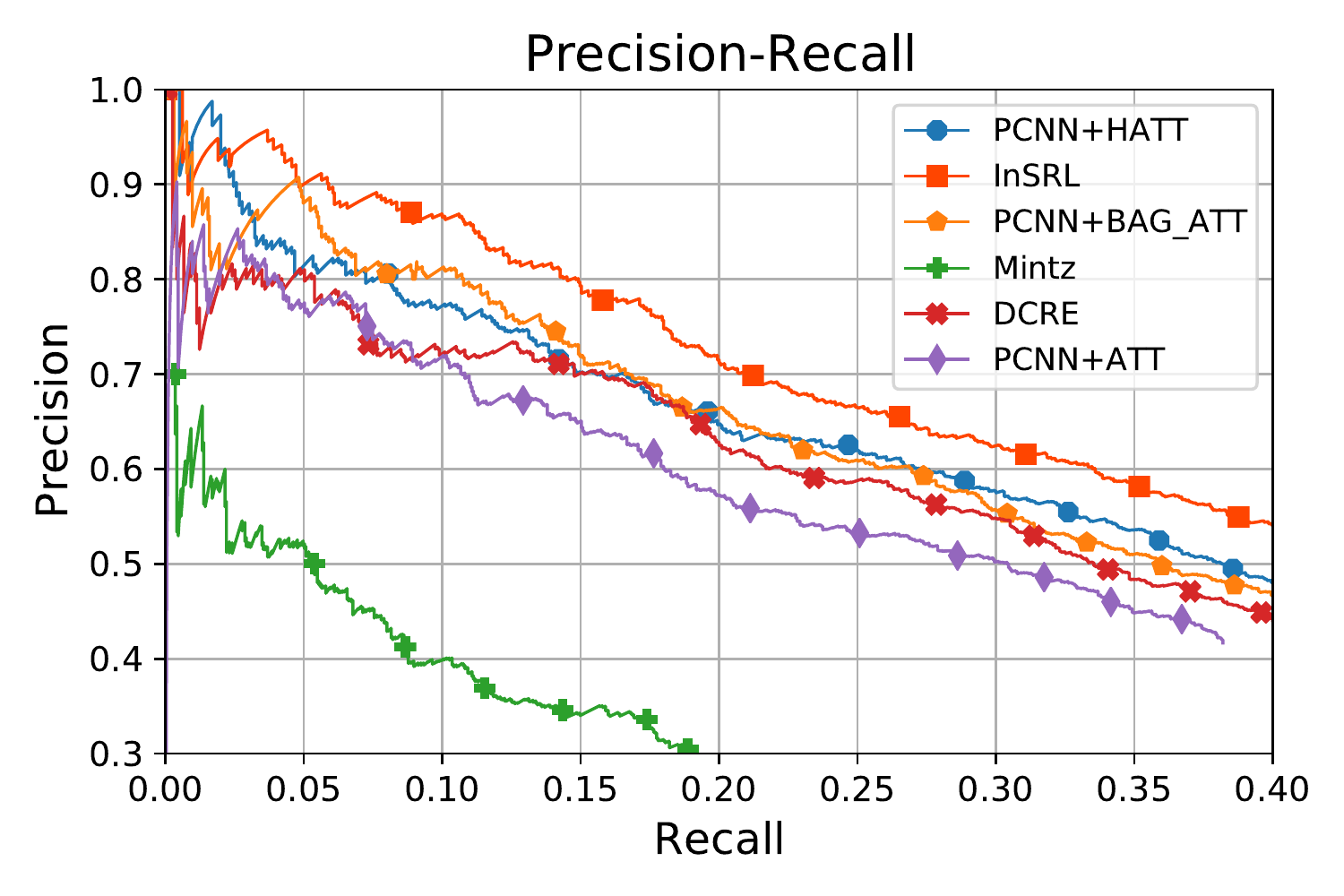}
    \caption{}
    \label{fig:my_label}
\end{subfigure}
\begin{subfigure}{0.45\textwidth}
\centering
    \includegraphics[width=7cm]{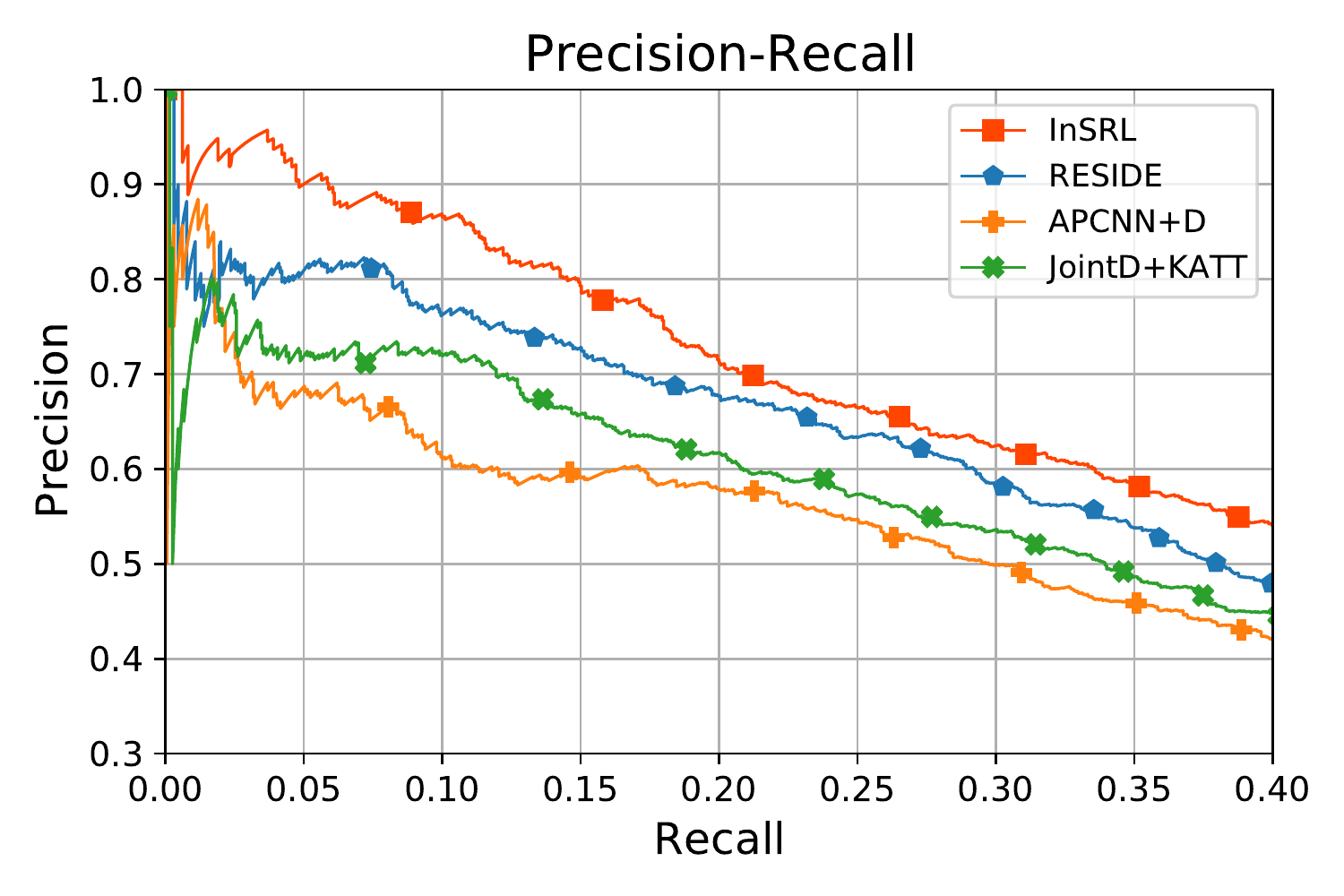}
    \caption{}
    \label{fig:my_label}
\end{subfigure}
\caption{Precision-recall curves comparison with (a) single-source based methods; (b) multi-source based methods.}
\label{fig:overall_comparison}
\vspace{-1em}
\end{figure*}

\subsection{Overall Evaluation}
\subsubsection{Baselines}
We consider several state-of-the-art baselines in our experiments, which can be divided into two categories: single source based and multi-source based methods.

(1) Single-source based methods. Mintz \cite{mintz2009distant}, a multi-class logistic regression model under distant supervision. 
PCNN+ATT \cite{lin2016neural} is a multi-instance learning model using piecewise max-pooling over CNN for sentence learning, where the sentence-level attention is conducted. PCNN+HATT \cite{han2018hierarchical} adopts hierarchical attention to utilize correlations among relations.
PCNN+BAG-ATT \cite{ye2019distant} uses intra-bag attention to deal with the sentence-level noise and inter-bag attention to handle the bag-level noise. DCRE \cite{shang2020noisy} utilizes unsupervised deep clustering to generate reliable labels for noisy sentences, which can further provide classifiable information.

 (2) Multi-source based methods. APCNN+D \cite{ji2017distant} uses entity descriptions to refine entity representation learning. JointD+KATT \cite{han2018neural} jointly conducts knowledge graph embedding learning and relation extraction. RESIDE \cite{vashishth-etal-2018-reside} consider entity type and relation alias information for imposing soft constraints while predicting relations.

\subsubsection{Results and Analysis}

We present the Precision-recall curves of all models in Figure \ref{fig:overall_comparison}. Overall, our InSRL achieves higher precision over almost the entire recall range.

(1) Comparison with single-source based methods.
All single-source based methods suffer from the \emph{noisy} and \emph{sparse} bag issues. PCNN+BAG\_ATT construct \emph{super-bag} to utilize useful information among similar bags. DCRE assign pseudo labels to noisy sentences rather than drop them. Both of them utilize a wider range of information, however, the constructions of super-bag and pseudo labels are in an unsupervised manner, which are more likely to amplify the influence of noisy sentences. PCNN+HATT employs hierarchical information of relations to better identify valid sentences but cannot help bags with only one sentence. In contrast, beyond the information of sentence bags, our InSRL exploits additional information sources. The superior performance performance proves they can effectively relieve the issues of sentence bags.

(2) Comparison with multi-source based methods.
APCNN+D proves the effectiveness of entity descriptions. However, they are only used to complement the entity representation learning, which ignores the potential classifiable information in descriptions. JointD+KATT jointly trains a knowledge graph completion and a RE model connected by mutual attention, so the information in knowledge graphs is still not explicitly introduced to RE. RESIDE directly includes entity types and relation alias information when training the RE model. However, entity descriptions are not considered. Besides, embeddings from different information sources are directly concatenated in RESIDE, which ignores the various importance of sources. Our InSRL constructs intact space embeddings for every entity pair by utilizing multiple information sources, which can keep useful information as much as possible. The results show that InSRL is effective to fuse multiple sources.


\subsection{Ablation study}

\subsubsection{Necessity of Multiple Sources}
The overall evaluation has shown that the introduced multiple sources can improve classification performance.
Furthermore, we study the impact of each source on the performance improvement.
Specifically, instead of considering all the three sources, we only take one or two sources for experiments (both training and testing). 
We present the results in Table \ref{View Combinations}.

We conclude that the results based on a single view is not competitive in performance, which indicates single view is not sufficient for RE. Besides, the results with labeled sentences (i.e., $v_1$) are better than those without  $v_1$, which demonstrates the labeled sentences are more informative than the other two views.
Moreover, entity descriptions (i.e., $v_2$) generate better results compared with multi-grained types (i.e., $v_3$).
Because many entity descriptions contain the relation information for some entity pairs, which significantly helps the relation identification. 
For example, the description of \texttt{My Heart Will Go On} strongly indicates the target relation, as shown in Figure \ref{fig:multiview_example}. 

\begin{table}[!htbp] 
	\caption{The experiment results with only one or two sources used. 
		$v_1,v_2,v_3$ denote the labeled sentences, entity descriptions and multi-grained types, respectively.}
	\centering
	\scalebox{1}{
		\begin{tabular}{ccc}
			\Xhline{2\arrayrulewidth}
			Views  &AUC    & $\max$ F1 \\
			\hline
			w/ $v_1$ &0.392 {\small $\pm 0.012$} &0.424 {\small $\pm 0.017$}\\
			\hline
			 w/ $v_2$ &0.310 {\small $\pm 0.014$} &0.314 {\small $\pm 0.016$}\\
			\hline
		      w/ $v_3$ &0.298 {\small $\pm 0.022$} &0.288 {\small $\pm 0.019$}\\
			\hline
			 w/o $v_1$ &0.327 {\small $\pm 0.031$}&0.331 {\small $\pm 0.029$} \\
			\hline
			 w/o $v_2$ &0.402 {\small $\pm 0.014$}&0.415 {\small $\pm 0.016$}\\
			\hline
			 w/o $v_3$ &0.413 {\small $\pm 0.015$}&0.441 {\small $\pm 0.018$}\\
			\hline
			InSRL &0.451 {\small $\pm 0.011$}&0.465 {\small $\pm 0.013$}\\
			\Xhline{2\arrayrulewidth}
		\end{tabular}}
	\vspace{-1em}
	\label{View Combinations}
\end{table} 


%
%

\subsubsection{Effectiveness of Intact Space Learning}
In our framework, we conduct intact space learning with view-level attention to integrate the three views, thus obtaining the intact space embeddings for entity pairs.
To demonstrate the importance and effectiveness of this module, we also provide some alternatives and observe performance changes in relation classification.
Specifically, (1) MV-AVG is the baseline where the three view embeddings are averaged.
Then we conduct a nonlinear transformation with the averaged vector as input, thus obtaining the integrated embedding for an entity pair.
That is, 
\begin{equation}
\boldsymbol x= \tanh({1\over 3}\boldsymbol{W}_6 \sum_{i=j}^3 \boldsymbol{v}_j  + \boldsymbol b_6 )
\label{MV-ISL (Average)}
\end{equation}
where $\boldsymbol{W}_6 \in \mathbb{R}^{d_x \times d_{m}}$ and $\boldsymbol b_6 \in \mathbb{R}^{d_x}$ are parameters.
Then $\{\boldsymbol x\}$ are used to construct the relation classifier using Eq \eqref{Softmax layer as follows} and \ref{function for RC task is}.
Note that, different from the intact space learning, we do not need to learn $\{\boldsymbol x\}$ and the relation classifier separately.
Instead, all the parameters in MV-AVG can be jointly learned within one loss function, i.e., Eq \eqref{function for RC task is}.
(2) MV-ATT is the improved version of MV-AVG.
In this baseline, instead of using simple average on all the views, we take rel-attention (i.e., Eq \eqref{attention weight of the RC task to view}) to compute the importance weight of each view, thus obtaining the weighted average vector.
Similar to MV-AVG, this vector will also be input to a nonlinear function, thus obtaining the embedding of an entity pair.
All the parameters will be jointly learned within one loss function. (3) InSRL (w/o RAT), where inner-view attention is removed, embeddings of words/types are directly averged to obtain the sentence/type set embeddings. (4) InSRL-AVG is the model where the intact space learning is still conducted but the cross-view attention is removed.
Thus, when constructing intact space representations, each view is equally important, i.e., $\gamma_{j}=1/3$ ($j=1,2,3$) in Eq \eqref{loss function single sample}.
We present the results in Table \ref{Necessity of Latent Embeddings for RC task}.

All variants of InSRL outperform both MV-AVG and MV-ATT, which indicates the ISL framework is effective to learn informative embeddings, this aligns with the theoretical analysis of stability and generalization ability of ISL.
However, these abilities cannot be guaranteed in both MV-AVG and MV-ATT. The comparison among InSRL (w/o RAT), InSRL-AVG and InSRL shows the effectiveness of our inner- and cross-view attention mechanisms.
\begin{table}[!htbp]
	\caption{The results with various fusion strategies.}
	\centering
	\scalebox{1}{
		\begin{tabular}{cccccccccc}
			\Xhline{2\arrayrulewidth}
			Model  &AUC    & $\max$ F1 \\
			\hline
			MV-AVG  &0.371{\small $\pm 0.021$} &0.392 {\small $\pm 0.023$}\\
			\hline
			MV-ATT  &0.413{\small $\pm 0.019$} &0.430{\small $\pm 0.024$}  \\
			\hline
			InSRL (w/o RAT)  &0.422 {\small $\pm 0.014$} &0.437 {\small $\pm 0.015$} \\
			\hline
			InSRL-AVG    &0.430 {\small $\pm 0.017$} &0.450 {\small $\pm 0.014$} \\
			\hline
			InSRL    &0.451 {\small $\pm 0.011$} &0.465 {\small $\pm 0.013$} \\
			\Xhline{2\arrayrulewidth}
		\end{tabular}
	}
	\label{Necessity of Latent Embeddings for RC task}
	\vspace{-2em}
\end{table}

\section{Conclusion \& Future works}
In this paper, we focus on the relation extract problem under the entity pair level. We analyze that only the sentence-based multi-instance learning is not sufficient for high-performance RE due to the \emph{sparse} and \emph{noisy} bags issues. Thus, we introduce multiple sources to provide complementary information. For this purpose, we present an end-to-end multi-view solution to effectively fuse information from various sources. Experiments show the necessity of multiple sources and effectiveness of our framework. Beyond the RE task, there are other scenarios where the fusion of information sources are needed, e.g., multi-modal learning. We plan to apply our framework to other multi-modal NLP tasks.

 \bibliography{aaai21}

\end{document}